# Neuroevolution-Enhanced Multi-Objective Optimization for Mixed-Precision Quantization

Santiago Miret[*][†]
santiago.miret@intel.com
Intel Labs, San Francisco, CA, USA

Vui Seng Chua[*]
vui.seng.chua@intel.com
Intel Labs, Hillsboro, OR, USA

Mattias Marder
mattias.marder@intel.com
Intel Datacenter and AI, Haifa, Israel

Mariano Phiellip
mariano.j.phielipp@intel.com
Intel Labs, Phoenix, AZ, USA

Nilesh Jain
nilesh.jain@intel.com
Intel Labs, Hillsboro, OR, USA

Somdeb Majumdar
somdeb.majumdar@intel.com
Intel Labs, San Diego, CA, USA

## ABSTRACT

Mixed-precision quantization is a powerful tool to enable memory and compute savings of neural network workloads by deploying different sets of bit-width precisions on separate compute operations. In this work, we present a flexible and scalable framework for automated mixed-precision quantization that concurrently optimizes task performance, memory compression, and compute savings through multi-objective evolutionary computing. Our framework centers on Neuroevolution-Enhanced Multi-Objective Optimization (NEMO), a novel search method, which combines established search methods with the representational power of neural networks. Within NEMO, the population is divided into structurally distinct sub-populations, or species, which jointly create the Pareto frontier of solutions for the multi-objective problem. At each generation, species perform separate mutation and crossover operations, and are re-sized in proportion to the goodness of their contribution to the Pareto frontier. In our experiments, we define a graph-based representation to describe the underlying workload, enabling us to deploy graph neural networks trained by NEMO via neuroevolution, to find Pareto optimal configurations for MobileNet-V2, ResNet50 and ResNeXt-101-32x8d. Compared to the state-of-the-art, we achieve competitive results on memory compression and superior results for compute compression. Further analysis reveals that the graph representation and the species-based approach employed by NEMO are critical to finding optimal solutions.

## 1 INTRODUCTION

Recent advances in deep learning have partially been driven by rapid growth in the size and complexity of deep neural network (DNN) architectures [5, 47]. This growth in complexity, along with a desire to deploy deep neural network workloads on various types of hardware, has spurred a significant need for advances in memory and compute compression techniques for DNN workloads. Memory compression refers to reduction in memory footprint required to store a given DNN on hardware, while compute compression refers to throughput acceleration achieved by performing optimization on the workload. In this study, we define memory compression as reduction in workload memory footprint compared to the original workload, and measure compute compression as the reduction of bit-operations during workload inference. The total bit-operations of a workload is defined as the number of Multiply-Add-Accumulation (MAC) operations multiplied by the compute precision. Bit-operations are a useful, hardware agnostic metric to measure latency and power consumption of workload, as indicated by studies which show that high-throughput low-precision compute devices consume lower power compared to floating-point units while also achieving lower latency [34]. Moreover, given that full-stack support for heavily quantized DNN model inference is limited, particularly in low precisions, hardware agnostic metrics such as memory footprint and bit-operations represent helpful proxies to compare results of different DNN compression studies with potentially different hardware systems.

In this work we focus on mixed-precision quantization [10, 15, 41, 44], a subset of quantization methods that reduce model weights and activations to a heterogeneous set of bit-widths throughout the model. The goal of bit-width reduction is to decrease a model's memory footprint and reduce latency at inference time with minimal degradation of task performance. Common approaches to mixed-precision quantization focus on optimizing one particular feature, usually the model memory footprint [10, 41]. Conversely, our approach, which is outlined in Figure 1, frames mixed-precision quantization as a multi-objective problem where we aim to find the Pareto optimal set of mixed-precision configurations that express the optimal trade-offs between various metrics of interest. In this work, we focus our multi-objective optimization approach on three distinct, objectives for each workload: task performance, model size (memory), and compute complexity (bit-operations).

We determine our multi-objective solutions through our novel method Neuroevolution-Enhanced Multi-Objective Optimization (NEMO), which combines classical multi-objective search algorithms [24] with neuroevolution methods that directly manipulate the parameters of a neural network using evolutionary operations [22, 39]. NEMO manages structurally distinct sub-populations, referred to as species, including neural network and non-neural network representations, via a dynamic allocation technique leveraging a multi-bandit approach and multi-objective utility functions.

As described in Section 4, we evaluate our method on various common ImageNet [8] workloads, which is common benchmark in image classification in machine learning. Based on our approach, we make the following contributions, which to best of our knowledge are novel contributions to the fields of neural network quantization and evolutionary computing:

(1) **NEMO**: a gradient-free, scalable, multi-objective evolutionary computing method enabling both traditional search

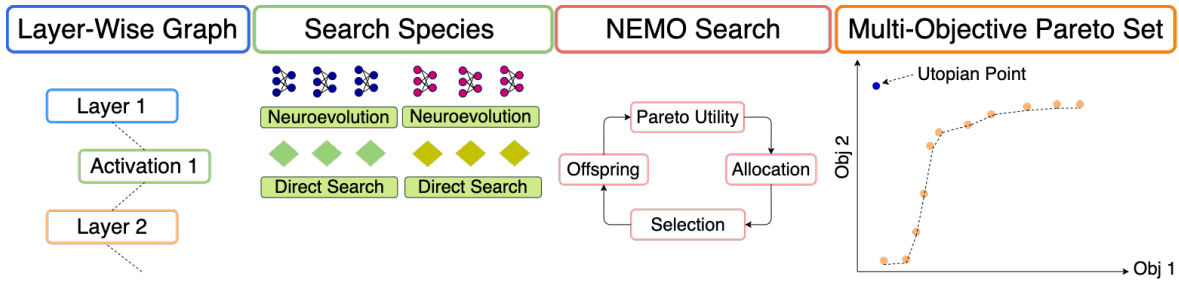

**Figure 1: Major Steps: First we refactor a workload into a layer-wise graph; second we generate distinct search species based on neuroevolution and direct search methods, third apply NEMO to search for a Pareto optimal set in multiple dimensions; last we report the Pareto optimal set for various mixed-precision workloads**

methods and neuroevolution techniques through distinct species in its population. NEMO improves on established search methods, such as NSGA-III [6], by dynamically managing structurally distinct species in its population based on the quality of their contribution to overall Pareto frontier. In this work, in addition to finding viable solutions leveraging species inspired by traditional search methods, NEMO successfully trains GNN representations to find Pareto optimal mixed-precision configurations using neuroevolution. Our results also show that NEMO can scale to the large combinatorial search spaces of real-world problems like mixed precision quantization with our largest workload, ResNeXt-101-32x8d [45] having a search space of $\sim 10^{206}$.

(2) **Graph Representation of Neural Network Workloads:**
We develop a graph-based representation of the neural network workloads allowing us to leverage graph neural networks trained by neuroevolution techniques to find suitable mixed-precision configurations for the weights and activations of the compressed workloads.

(3) **Multi-Objective Formulation for Mixed-Precision Quantization:** We create a multi-objective framework for mixed-precision quantization leading to a Pareto optimal set of solutions expressing the trade-offs related to various metrics relevant for model compression. This multi-objective framework provides a more general solution to the mixed-precision quantization problem and can be used to find optimal configurations for various conditions, which alleviates the need for recurring searches often needed in single objective approaches.

## 2 RELATED WORK

**Multi-Objective Search:** Our mixed-precision quantization technique is enabled by NEMO, which differentiates from traditional multi-objective evolutionary search algorithms [24] and past neuroevolution techniques, such as MM-NEAT [30, 36, 42], and more recent algorithmic variations [28, 38] in a couple of meaningful ways: NEMO enables dynamic species management between structurally distinctive species, including GNN based species and model free species that search directly in the solution space. This enables architectural diversity across the population, including distinct mutation and crossover operations for each species, to operate together in a unified framework. Prior work [2, 37], conversely, usually applied similar mutation and crossover operations across species to achieve diversity in the solution space. Moreover, NEMO leverages multi-objective utility metrics measured for each species with multi-bandit methods in an integrated framework that can be applied to large variety of search problems beyond the mixed-precision quantization setting described in this work.

**Quantization:** Deep neural network compression via quantization has become increasingly prevalent in recent years, as evident by INT8 quantization support by major deep learning frameworks [1, 33], as well as hardware vendors enabling developers to perform efficient model quantization. The research community has continued to push the state of the art in quantization methods through various approaches, including quantization-aware training [11, 19], data-free quantization methods [17, 31] , as well as post-training [29] and mixed-precision quantization. Recent mixed-precision quantization research, such as DQ [41], HAQ [44] and HMQ [15], has generally focused on solving a constrained optimization problem where memory compression is optimized while maintaining a minimal, acceptable accuracy drop. Within the aforementioned constrained optimization settings, HAQ applied reinforcement learning, while DQ and HMQ applied gradient based optimization to a differentiable formulation of bit-width selection. Further advances, such as HAWQ-V3 from Yao et al. [46] and HMQ from Habi et al. [15] developed mixed-precision quantization techniques considering hardware-aware metrics for further optimization. Yao et al. [46] created a hardware-aware mixed-precision quantization method leveraging integer linear programming and measured bit-operations as a surrogate metric for hardware performance. Habi et al. [15] developed a mixed-precision quantization block, which is applied to the workload to find uniform and symmetric bit-widths for the given block. Habi et al. [15] also reported Pareto frontiers for Cifar-10 [27] and ImageNet [8] workloads in two dimensions, accuracy and memory compression. Our work, by contrast, reports a Pareto frontier in three relevant dimensions including accuracy, memory compression and bit-operations as a surrogate metric for hardware performance, and can be easily extended to more objectives depending on the broader task or specific user needs. We also develop a graph representation of the workload inspired by Khadka et al. [21], allowing us to leverage graph convolutions trained by gradient-free neuroevolution, which to the best of our knowledge is a novel approach to mixed-precision quantization.

## 3 METHOD

Our method is based on the integration of three techniques: 1. Formulating layer-wise mixed-precision quantization of a workload as a multi-objective search problem enabling us to apply traditional search methods as distinct species. 2. Refactoring of the workload into a sequential graph representation suitable for GNNs, which we can formulate as neuroevolution species. 3. Finding a Pareto optimal set of solutions using NEMO leveraging traditional search and GNNs as separate species in the population. We create our graph representation by declaring each quantizable operation in the workload as a node and build the edges by connecting the nodes sequentially, meaning $\mathcal{G}_{\text{edges}} = \{n_1 \to n_2, n_2 \to n_3, \ldots\}$, where $n$ corresponds to the graph nodes representing computational operations in the targeted workload. The node features of the graph representation consist of a concatenation of a one-hot encoding of the given operation and four general features associated with the described in detail in Appendix B.2. After constructing the workload graph, we perform inference on the nodes of the graph either by direct evolutionary search or with a graph neural network as shown in Figure 2. Each inference method, including different GNN architectures and different traditional search methods, is codified as a distinct species within NEMO.

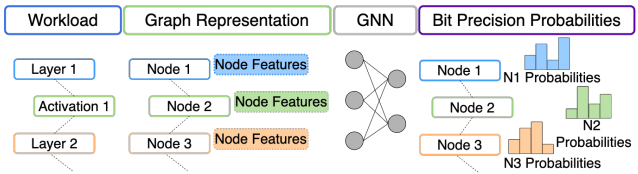

Figure 2: Bit-Precision Selection For A Workload: 1. We transform the workload into a sequential graph; 2. We perform an inference method of choice on the graph (direct search on the bit-widths or Graph Neural Network inference); 3. We select the bit-width based on the resulting probabilities.

### 3.1 Mixed-Precision Quantization

We perform uniform affine quantization, the predominant approach for quantized neural network inference [13, 26]. Affine quantization maps neural network weight and activation tensors of 32-bit floating point precision to a finite set of fixed points defined by the quantization bit width. Specifically, we employ asymmetric affine quantization to neural network layers, where each weight (parameter) and activation (input) tensor of a layer is put through a quantization function $q$ defined as:

$$q(\boldsymbol{x}; \boldsymbol{b}) = \left\lfloor \frac{clamp(x; x_{min}, x_{max})}{s} + z \right\rceil;$$
$$s = \frac{x_{max} - x_{min}}{2^b - 1}; \ z = \frac{-x_{min}}{s} \quad (1)$$

Each quantizer $q$ maps the elements $x \in \mathbb{R}$ of a tensor $\boldsymbol{x}$ to a quantized value corresponding to one of the integers $\{-2^{b-1}, -2^{b-1} + 1, -2^{b-1}+2, \ldots, 2^{b-1}-1\}$, and is parameterized by the bit width $b$, as well as $x_{min}, x_{max}$ corresponding to the thresholds of the dynamic range of the tensor $\boldsymbol{x}$ which are calibrated based on the elements of tensor. As shown in Equation (1), $\{b, x_{min}, x_{max}\}$ collectively define the scale factor $s$ and zero point $z$. $s$ denotes the step size in $x$ representing a unit delta between two adjacent integers, and $z$ corresponds to the value in $x$ which maps to the zero quantized value. Effectively, each quantizer $q$ first saturates the tensor $\boldsymbol{x}$ by $clamp(x; x_{min}, x_{max}) := min(max(x; x_{min}), x_{max})$, then scales the saturated tensor by $s$, adds the offset $z$ and finally rounds the resulting tensor elements to the nearest integer. Effective quantization can reduce the memory footprint of a workload and also increase inference speed by enabling efficient usage of high throughput low-precision computational units. These benefits of quantization, however, usually come at the cost of lower performance of the neural network on its desired task. Given these inherent trade-offs between task performance, memory footprint and computational throughput, we formulate finding fine-grained quantizer bit-widths as a multi-objective search problem. Given an $L$-layer trained neural network $N$, we search for the set of bit widths $\mathcal{B}$ consisting of weights ($b_w^l$) and activations ($b_a^l$) for each layer subject to $K$ objectives:

$$\mathcal{B} = \{b_w^0, b_a^0, b_w^1, b_a^1, \ldots, b_w^l, b_a^l\} \ \forall l \in L$$
$$\text{s.t.} \ \{obj_0(N'), obj_1(N'), \ldots, obj_k(N')\} \ \forall k \in K$$

As shown in greater detail in Figure 2 and Figure 3, we obtain a Pareto optimal of mixed-precision configuration $\mathcal{B}$ by leveraging a combination of direct evolutionary search and graph neural network inferences, both of which are trained and refined by our NEMO method.

### 3.2 Neuroevolution-Enhanced Multi-Objective Optimization

Neuroevolution-Enhanced Multi-Objective Optimization (NEMO), shown in Figure 3 and outlined Algorithm 1, is a multi-objective evolutionary search algorithm with the capability of managing multiple species. Many of the components of the NEMO algorithm can be tailored towards different needs and preferences, e.g. species types may contain various architectures and utility metrics can express any underlying property one aims to optimize. As described in Algorithm 1, NEMO manages both species level and individual levels variables. Species level variables include the utility metric $u_{s \in \mathcal{S}}$, the species allocation $a_{s \in \mathcal{S}}$, as well as the mutation and crossover operations, $\mathcal{M}_{s \in \mathcal{S}}$ and $C_{s \in \mathcal{S}}$. Individual level variables, which are tracked for each individual in the population regardless of the species, include the fitness $\mathcal{F}_{p \in \mathcal{P}}$ and the rank $r_{p \in \mathcal{P}}$.

In this study, we specify the R2-Indicator [16], with a set of uniform weight vectors $\Lambda$, as the utility metric for each species $u_{s \in \mathcal{S}}$. The R2 corresponds to the averaged sum of the minimum distances for all solutions $\gamma \in \Gamma$, where $\gamma$ is analogous to $\mathcal{F}_{p \in \mathcal{P}}$ above, in any dimension over each weight vector $\lambda \in \Lambda$:

$$R2(\Gamma) = R2(\Gamma, \Lambda, z^*)$$
$$= \frac{1}{|\Lambda|} \sum_{\lambda \in \Lambda} \min_{\gamma \in \Gamma} \left\{ \max_{i \in 1, \ldots, k} \{ \lambda_i | z_i^* - \gamma_i | \} \right\} \quad (2)$$

$z^*$ above corresponds to an utopian point that represents the best possible solution (in the case of mixed-precision quantization the

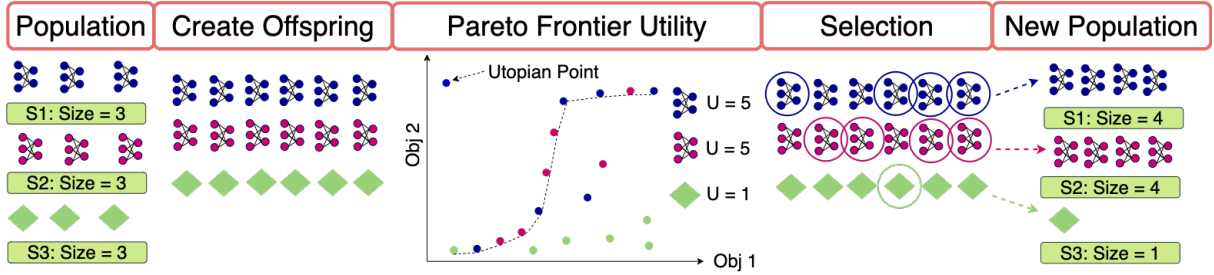

**Figure 3: An example NEMO generation for a population $\mathcal{P}$ with $\mathcal{P}_{size} = 9$ containing three species $s \in \mathcal{S}$ of equal size $s_{size} = 3$: {Neuroevolution Species #1 (S1 - blue networks), Neuroevolution Species #2 (S2 - magenta networks) and Search Species (S3 - green diamond)}. At the beginning of a generation each species produces offspring, doubling the number of members in each species. NEMO then computes the utility metrics for each species (in the illustration above the utility is the number of solutions in the Pareto frontier), as well as each species' respective size allocations. Finally, the best-performing members within each species' size allocations are selected for the next generation.**

---

**Algorithm 1** Generalized NEMO Algorithm

1:  Initialize a population $\mathcal{P}$ with individuals $p$ contained in a set of species $p \in s \in \mathcal{S} \in \mathcal{P}$
2:  Define species specific mutation operation $\mathcal{M}_{s \in \mathcal{S}}$ and crossover operation $C_{s \in \mathcal{S}}$
3:  Initialize an allocation method $\mathcal{A}_{\mathcal{P}}$, a utility function $\mathcal{U}_{\mathcal{P}}$ describing the utility of each species $s \in \mathcal{S}$ and a ranking method $\mathcal{R}_{\mathcal{P}}$ for the population $\mathcal{P}$
4:  Define stopping criteria (e.g. maximum number of generations)
5:  Evaluate the fitness $\mathcal{F}_{p \in \mathcal{P}}$ for each member of the population (e.g [accuracy, memory, bit-ops])
6:  **while** training NEMO **do**
7:      **for** each generation **do**
8:          Apply mutations $\mathcal{M}_{s \in \mathcal{S}}$ and crossovers $C_{s \in \mathcal{S}}$ to create offspring $O_{s \in \mathcal{S}}$
9:          Evaluate the fitness $\mathcal{F}_{p \in O}$ for each member in the offspring
10:         Compute the utility $u_{s \in \mathcal{S}}$ for each species using $\mathcal{U}$
11:         Compute the allocation $a_{s \in \mathcal{S}}$ for each species using $\mathcal{A}$
12:         Compute the rank $r_{p \in \mathcal{P}}$ for each member of population using $\mathcal{R}$
13:         **for** $s \in \mathcal{S}$ **do**
14:             Fill $a_{s \in \mathcal{S}}$ starting with the highest ranked members of the species $p \in s \in \mathcal{P}$
15:             **if** $s_{size} < a_{s \in \mathcal{S}}$ **then**
16:                 Select random offspring until $s_{size} = a_{s \in \mathcal{S}}$

utopian point represents 100% accuracy, 0 bytes model-size and 0 bit-ops). As discussed in Künzel and Meyer-Nieberg [28], Steven [38], minimizing the R2 indicator for a given set of solutions incentivizes the algorithm to find solution closer to the utopian point $z^*$, which is a good fit our purposes. Another potentially promising choice of utility metric is the hypervolume spanned by a given Pareto front [28, 38]. Our setup with multiple species, however, can easily lead to overlapping hypervolumes making it more challenging to attribute the set-wise contribution of a species compared to the R2-Indicator. To determine species size allocation we compute an upper confidence bound (UCB) score $\mathcal{A}_{s \in \mathcal{S}}$ [3] commonly applied

in multi-bandit problems [4, 20], which is given by:

$$\mathcal{A}_{s \in \mathcal{S}} = v_s + c * \sqrt{\frac{\log(\sum_{s \in \mathcal{S}} y_s)}{y_s}} \qquad (3)$$

where $v_s$ is the utility (R2 value) and $y_s$ is the number of evaluations of the species respectively. For ranking and selection we apply a variant of NSGA-III by Deb and Jain [6], where we rank the individuals in the entire population based on their multi-objective solution vectors, and then fill the size allocation of each species based on the overall NSGA-III population ranking. If any species has leftover allocation after the filling process, we select an additional set of random offspring to fill its remainder and continue to the next generation.

## 3.3 Search Species

Within the NEMO algorithmic framework shown in Algorithm 1, we define a set of species that perform our search:

- **Continuous Direct Search:** We define a continuous action space between the lowest and the highest bit-width precision, and round to the nearest integer within the set of possible precisions when selecting the final bit-width for each layer. Here, we leverage Polynomial-Bounded mutation and Simulated-Binary-Bounded crossover from Deb et al. [7] to create offspring within the species in each generation.

- **Floor-Rounding Direct Search:** We define the choice of bit-widths as integers within the space of possible bit-width choices. Here, we also leverage Polynomial-Bounded mutation and Simulated-Binary-Bounded crossover from Deb et al. [7] to create offspring within the species. Given that the mutation and crossover happen in continuous space, we round to integer precision by choosing the integer that is nearest to the continuous parameter, but also lower in value. This effectively biases the rounding to go towards lower bit-width values, incentivizing more compression compared to *Continuous Direct Search* described above.

- **Sequential Graph - GCN:** We construct a GNN architecture similar to GCN-Conv by Kipf and Welling [23] augmented with graph-attention layers [43] that processes a sequential graph described in Section 3. The mutation operation is based on sub-structure neuroevolution (SSNE) [22, 40], which is described in Algorithm 2. Here, we loop through each tensor inside the neural network individuals of the species and perform sub-structured based mutation and uniform crossover operations with probabilities $mut_{dist}$ and $cr_{dist}$, respectively. The bit-width values are chosen according to the logits output of GNN, which represent probabilities for each bit-width as shown in Figure 2.
- **Sequential Graph - Graph-UNet:** We construct a GNN architecture similar to Graph U-Net by Gao and Ji [12] augmented with graph-attention layers [43] that processes a sequential graph described in Section 3. Similar to the *Sequential Graph - GCN* species, we leverage SSNE for mutation and crossover operations and choose the bit-widths values according to the logits of the GNN.

The general structure of SSNE applied within NEMO, specifically to the *Sequential Graph - GCN* and *Sequential Graph - Graph-UNet* species is shown in Algorithm 2:

---

**Algorithm 2** NEMO Species Sub-Structure Neuroevolution

---

1: Define a random number generator $r()$ with output $\in [0, 1)$
2: Define crossover and mutation probabilities $cr_{dist}$ & $mut_{dist}$
3: Define mutation strength $mut_{strength}$
4: Extract neuroevolution species $s_t \in \mathcal{S}_t$ from NEMO population $\mathcal{P}_t$ at iteration $t$
5: **for** Neuroevolution species $s_t \in \mathcal{S}_t$ with $k$ individuals **do**
6:   **for** each crossover iteration $k/2$ **do**
7:     Randomly extract two neural networks $N_i$ & $N_j \in s_t$
8:     **if** $r > cr_{dist}$ **then**
9:       **for** tensor substructure $T_i$ & $T_j \in N_i$ & $N_j$ **do**
10:         Perform uniform crossover of $\{T_i, T_j\}$ with probability $cr_{dist}$
11:   **for** each mutation iteration $k$ **do**
12:     Randomly extract a neural networks $N_i \in s_t$
13:     **if** $r > mut_{dist}$ **then**
14:       **for** tensor substructure $T_i \in N_i$ **do**
15:         Sample random index $k_j$ along each dimension of $T_i$ and mutate $T_i[k_j]$ by adding random noise from a Gaussian distribution $\mathcal{G}(\mu = 0, \sigma = mut_{strength} * T_i[k_j])$
16:     Place updated species $s_{t+1} \in \mathcal{S}_{t+1}$ in broader NEMO population $\mathcal{P}_{t+1}$ and follow ranking and selection procedures

---

## 4 EXPERIMENTS & RESULTS

We perform our experiments on different ImageNet workloads (MobileNet-V2 [35], ResNet50 [18], ResNeXt-101-32x8d [45]) based on pre-trained models provided in TorchVision [33], and leverage NNCF by Kozlov et al. [25] to automatically insert quantization layers for the weights and activations in the workload, including the first and last layers. While NNCF supports many quantization modes and settings, our experiments leverage asymmetric affine quantization to provide maximum flexibility for finding optimal configurations. NNCF initializes the dynamic threshold of each quantizer by calculating the statistical mean across a number of batches from the inference data. Following calibration, we extract relevant attributes of the NNCF representation of the quantized network as features of the graph structure described in our method section. We then allow our search algorithm to choose from a set of seven different bit-widths: {2, 3, 4, 5, 6, 7, 8} and perform a multi-objective optimization using NEMO to obtain a set of Pareto optimal precision maps based on the following objectives:

- **Task Performance:** max(*Top5 Accuracy*) of the given workload on a stratified subset of the ImageNet training data consisting of 5 images per class leading to a total of 5000 images. The stratified sample enables us to perform faster evaluations during the search process, given the significantly smaller dataloading required compared to > 1$M$ images in the full training dataset. We optimize Top5 accuracy to achieve less noisy evaluations during search and report results in Top1 accuracy in our results.
- **Memory Compression:** min(*Model-Ratio*) of a given workload that expresses the ratio of the reduced memory footprint of model parameters compared to the full-precision footprint.
- **Compute Compression:** min(*Bit-Operation Ratio*) of a given workload that outlines the ratio of reduced complexity compared to full-precision complexity.

NEMO then finds a three-dimensional Pareto frontier of optimal precision maps based on our stratified sampling of the ImageNet training data. At the conclusion of the search, we re-evaluate the Pareto set of configurations on the full ImageNet validation dataset of 50,000 images. Finally, we report our results with three-dimensional heatmaps containing all objectives, two-dimensional projections of Task Performance vs. Memory Compression, as well as two-dimensional projections Task Performance vs. Compute Compression.

Throughout the search process, we take advantage of the parallelizability of independent evaluations for each member of the population and engineer a parallel computation framework to concurrently evaluate the fitness of individuals within each of our search species. Once the fitness values for all individuals within a species are determined, we proceed to the next species and repeat our parallel evaluations. Overall, this parallel-based computation orchestration enables to significantly reduce the wall-clock time of our experiments as described in more detail in Appendix A.

### 4.1 Multi-Objective Pareto Optimal Sets

The combinatorial complexity of each experiments increases exponentially with the number quantizable operations for each workload. MobileNet-V2 contains 116 quantizable operations, leading to an combinatorial complexity of $7^{117}$ or $\sim 10^{98}$; ResNet50 includes 125 quantizable operations ($\sim 10^{105}$), and ResNeXt-101-32x8d contains 244 quantizable operations with a complexity of $\sim 10^{206}$. Our reported results pertain to quantized models evaluated on the full validation set of ImageNet. This part of the overall dataset does not include images used for the stratified sub-sampling used during the search process. This type of data splitting is common practice in machine learning and provides greater confidence that the quantized models we obtain are not overfit to the data used for fitness evaluations during the NEMO search. The uncompressed Top1 validation accuracies for MobileNet-V2, ResNet50 and ResNeXt-101-32x-8d are 71.878, 76.130, 79.312 respectively.

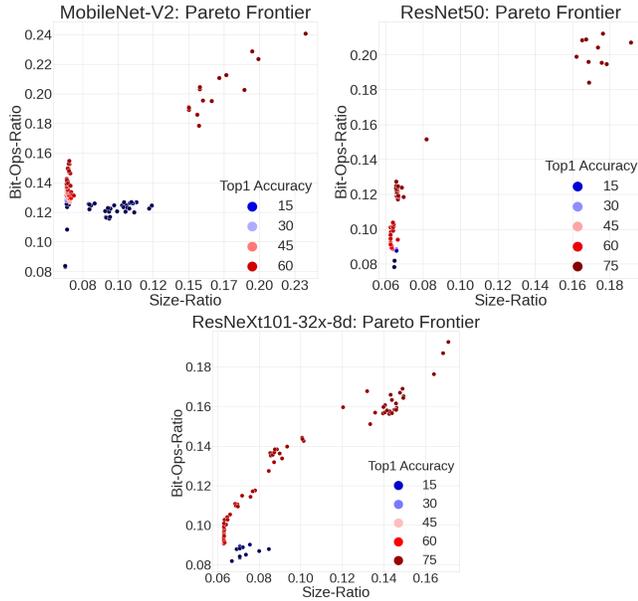

**Figure 4: Pareto Optimal Frontier Heatmap with all Three Objectives for MobileNet-V2, ResNet50 and ResNeXt-101-32x8d.**

Figure 4 shows the heatmaps for all workloads in our experiments. We generally observe a sparse distribution of Pareto Optimal solutions for all three workloads, most of which tend to have higher Top1 accuracy as seen by the prevalence of red-colored solutions in Figure 4. This suggests that mixed-precision quantization tends to have steep and significant changes in task performance in critical regions of the solution space. In other words, there is a compression level up to which task performance drop is generally minimal and beyond which task performance degradation is significant.

This trend is also observed in the two-dimensional Pareto frontier projections for memory compression in Figure 5 and compute compression in Figure 6. Here we can see the critical regions of steep Top1 Accuracy drop more clearly. In the case of compute compression, the threshold for steep task performance degradation lies slightly below a model-ratio of 0.075, which represents ∼ 13× compression in memory footprint. We also observe that for model-ratios > 0.075, ResNet50 has generally the lowest Top1 Accuracy drop, while ResNeXt-101-32x-8d has the largest. The difference between ResNet50 and ResNeXt-101-32x-8d can likely be attributed to the larger combinatorial search space of ResNeXt-101-32x-8d, which makes it significantly more challenging to find more optimal solution when compared to ResNet50. We also observe that the performance drop for MobileNet-V2 is generally larger compared to the other workloads, which we believe is related to MobileNet-V2's generally inferior performance on ImageNet compared to the other workloads. The lower performance of MobileNet-V2, however, is offset by its smaller absolute memory footprint. In cases where total memory footprint is more important, such as cell phones or other endpoint devices, MobileNet-V2 continues to be a better fit, while ResNet50 and ResNeXt-101-32x-8d are a better choice where task performance is more important, such as data centers and cloud computing.

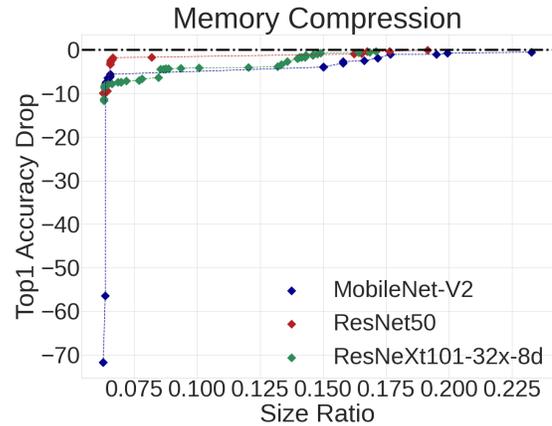

**Figure 5: Memory Compression: Two-Dimensional Projection of Pareto Optimal Frontiers for Memory Compression, expressed as Model-Ratio, for MobileNet-V2, ResNet50 and ResNeXt-101-32x-8d.**

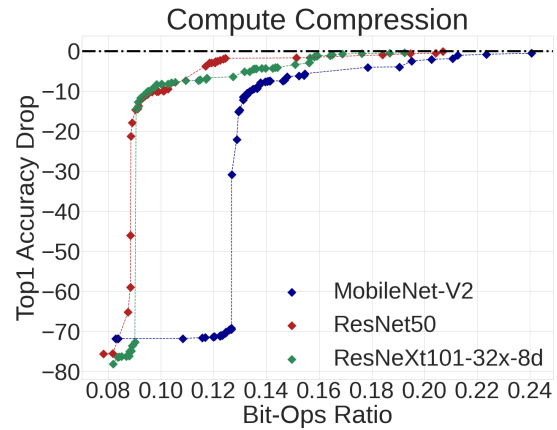

**Figure 6: Compute Compression: Two-Dimensional Projection of Pareto Optimal Frontiers for Compute Compression, expressed as Bit-Ops-Ratio, for MobileNet-V2, ResNet50 and ResNeXt-101-32x-8d.**

In the case of compute compression, we observe a generally similar behavior with steep task performance degradation beyond a given threshold. In this case, however, the threshold for MobileNet-V2 (bit-ops ratio: ∼ 0.13, compute compression: ∼ 8×), differs from the thresholds of ResNet50 and ResNeXt-101-32x-8d (bit-ops ratio:

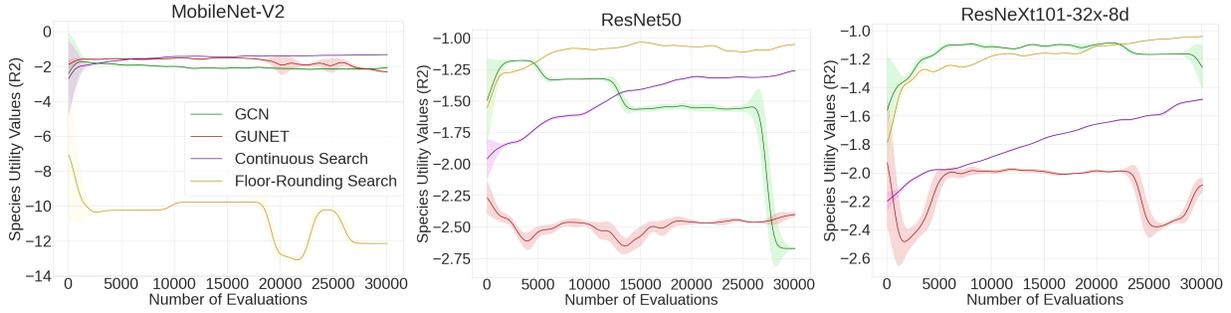

**Figure 7: R2 Utility Values for Different Species in NEMO Affecting Size of Each Species**

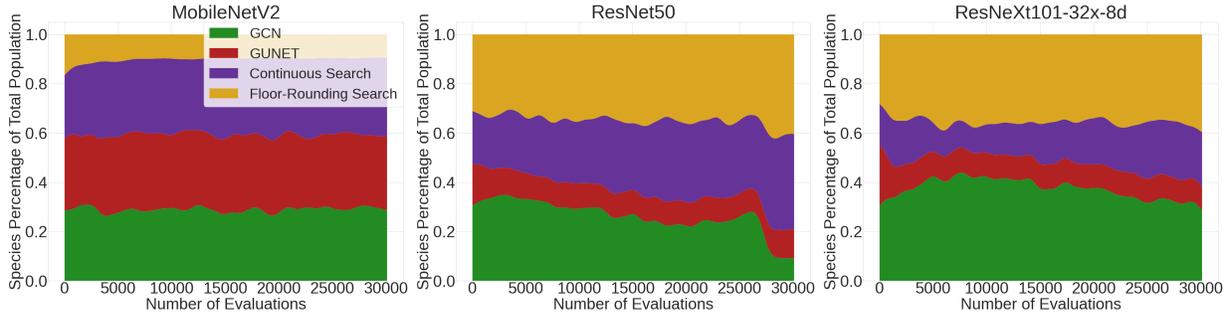

**Figure 8: Size Allocation Values for Different Species in NEMO (Population Size = 50, Minimum Species Size = 5)**

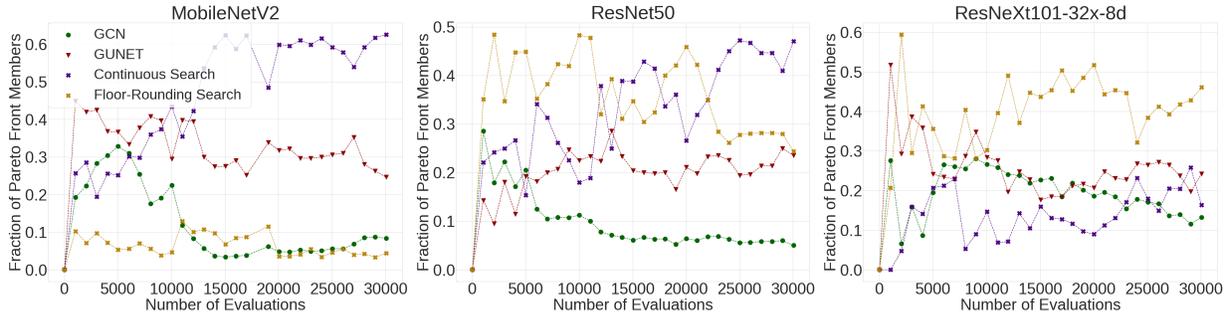

**Figure 9: Percentage of Members in the Pareto Optimal Frontier for Different Species in NEMO**

$\sim 0.09$, compute compression: $\sim 11\times$). This can also be attributed to MobileNet-V2's more compressed architecture compared to the other workloads, as there is less ability to reduce redundancy to further optimize compute compression.

## 4.2 NEMO Search Analysis

One of the major innovations of NEMO is the dynamic allocation of different pre-defined species in the search framework. As such, we wanted to assess the effects of having different species in the search method. Based on the R2 utility displayed in Figure 7, the species size allocation in Figure 8 and the percentage of Pareto frontier members for each of the species in Figure 9, we make the following observations:

- **Dynamic Mix of Species Contributions:** We observe that the mix of Pareto frontier members changes significantly for each workload. One example is that the Floor-Rounding species performs poorly on MobileNet-V2, both in terms of R2 utility and members in the Pareto frontier, while performing quite strongly for ResNeXt101-32x-8d. The Graph-UNet based species, on the other hand, performs well on MobileNet-V2 and generally poorly compared to the other species for ResNet50 and ResNeXt101-32x-8d. One of the important advantages of the dynamic species management of NEMO is that the algorithm can automatically adjust the allocation of species not only throughout the search, but also for different workloads, allowing the user to apply the same algorithm to various problems out-of-the-box.

- **Changing Species Utility During Search:** We observe that in some cases the utility of a given species can change depending on the quality of the solutions that particular species provides. This is particularly notable in the *Continuous Search* species for the ResNet50 and ResNeXt101-32x-8d workloads, where the utility, as shown in Figure 7, concurrently increases with the size of the species' allocation shown in Figure 8. This reinforces the beneficial ability of NEMO to adjust allocations based on the goodness of the species.

- **Correlated Species Allocation and Utility:** The data in Figure 7 and Figure 8 show that NEMO effectively allocates species size by the provided utility, which is the intended effect of the UCB score defined in Equation (3).

## 5 DISCUSSION

**Comparison to Quantization Literature:** Comparison of mixed-precision quantization results across different works of literature is generally very challenging, largely due to the fact that different engineering and design decisions in formulating up the optimization problem can lead to substantial differences in the final problem settings. As such, lateral comparisons between two different works require careful examination of how the authors of each study formulated and executed the optimization problem, and how relevant their method, metrics and reported results may be to the setting one is considering. Moreover, our case is further complicated by the fact that nearly all quantization literature focuses on single-objective optimization with regards to memory compression, compelling us to investigate whether we can convert the reported results to the metrics we studied.

| Method | Top1 Accuracy (%) | Memory Comp. | Compute Comp. | Search Space |
|---|---|---|---|---|
| MobileNet-V2 | 71.88 | $1\times$ | $1\times$ | |
| HAQ | 66.70 | $14.1\times$ | $1.0\times$ | $10^{52}$ |
| HMQ | 65.70 | $14.4\times$ | $4.0\times$ | $10^{49}$ |
| NEMO | 66.24 | $15.3\times$ | $6.5\times$ | $10^{98}$ |
| ResNet50 | 76.13 | $1\times$ | $1\times$ | |
| HAQ | 70.63 | $15.5\times$ | $1.0\times$ | $10^{56}$ |
| HMQ | 75.00 | $15.7\times$ | $4.0\times$ | $10^{52}$ |
| NEMO | 74.34 | $15.0\times$ | $8\times$ | $10^{105}$ |
| ResNext101-32x-8d | 79.312 | $1\times$ | $1\times$ | |
| NEMO | 74.92 | $11.5\times$ | $7.3\times$ | $10^{206}$ |

**Table 1: Comparison of NEMO in high-memory compression region for MobileNet-V2 (yellow) and ResNet50 (green). NEMO achieves better compute compression for all workloads. The last row represents ResNeXt101-32x-8d for which we found no additional results in the literature.**

With these caveats in mind, we outline how our solutions in corresponding neighborhoods of high memory compression compare to HAQ's k-means weight quantization [44] and HMQ's learned layer-wise weight bit width with uniform 8-bit activation [15]. As described in Section 2, HAQ leverages a reinforcement learning based approach to mixed-precision quantization, while HMQ applies gradient descent to optimize a differentiable formulation of the mixed-precision quantization challenge. Our results in Table 1 are based on reported Top1 Accuracy, compression ratio, and bit-ops for the high-memory compression region, as well as the size of the search space tackled for each formulation. Given that bit-ops were not reported in either paper, we infer the bit-ops based on the information about the final quantized workload as described in the respective papers to the best of our ability. We observe that the configurations found in the NEMO Pareto frontier are generally competitive in memory compression and state-of-the-art in compute compression across all three workloads we investigated. This is particularly noteworthy given that HAQ and HMQ leverage gradient based methods to fine-tune the parameters of their workloads after finding a given configuration of bit-widths. NEMO, by contrast, finds competitive and sometimes superior solutions entirely through search on the bit-widths space without any changes to the workload parameters. It is also worth mentioning that NEMO operates in significantly larger search spaces than HAQ and HMQ due to the greater flexibility of choosing the bit-widths for activation tensors. We believe the ability of NEMO to perform well, while searching in a greater combinatorial search space, highlights to promise of the method to deliver relevant solutions for real-world engineering problems.

**Future Work:** While our results in Section 4.1 show the promise of applying NEMO to real-world problems with large combinatorial search spaces, our method currently relies on an effective graph representation and pre-determined GNN architectures that act as species within NEMO. Future work could address this by discovering new species automatically, similar to an AutoML method or NEAT's historical markings [37].

On the quantization side, we only show results on ImageNet models, which are significantly smaller than language models, such as BERT [9], whose greater parameter space would make the search even more challenging. Moreoever, all results on mixed-precision quantization, including our own, are reported on simulation and not an actual hardware runs due to lack of hardware support for lower precision computations. We are encouraged by recent announcement from major hardware vendors [14, 32] that outline roadmaps for supporting mixed-precision quantization in future products. Once hardware support for lower integer precisions is available, one could apply our method on actual systems and investigate how it performs across a plethora of objectives that can only be measured in real systems, such as latency and power consumption. Beyond this study, we also see promise in applying our method to other neural network optimization challenges, such as compression via sparsification or hyperparameter optimization. Lastly, we also believe NEMO could be useful in multi-objective optimization challenges in different domains, especially those where different formulations of the problems can be represented by architecturally distinct species.

## A COMPUTATION DETAILS

The primary compute bottleneck for our search experiments was the evaluation of the simulated quantized network according to the procedure described in Section 3.1. The evaluation takes ~ 60 seconds for MobileNet-V2 and ResNet50, and ~ 100 seconds for ResNeXt101-32x-8d on a hardware unit with 1 GPU and 7 CPU cores as shown in Table 2.

| Evaluations | Evaluation Core Time | Number of Evaluations |
|---|---|---|
| MobileNetV2 | ~ 60s | 30,000 |
| ResNet50 | ~ 60s | 30,000 |
| ResNext101-32x-8d | ~ 100s | 30,000 |

**Table 2: NEMO Search Evaluation Details**

In order to accelarate the wall-clock time of our experiments, we parallelize the search with 10 parallel search instances and obtain the following approximate compute cost shown in Table 3 below.

| Compute Time | Core Time (Total) | Wall Clock (Total) |
|---|---|---|
| MobileNetV2 | 500 hours | 48 hours |
| ResNet50 | 500 hours | 48 hours |
| ResNext101-32x-8d | 850 hours | 72 hours |

**Table 3: NEMO Search Compute Time Details**

## B IMPLEMENTATION DETAILS

### B.1 NEMO Hyperparameters

Table 4 shows the hyperparameters applied for the NEMO algorithm for all our experiments. The search algorithm hyperparameters in Table 4 are separate from hyperparameters used to define each species, such as the specific architectures of the GNNs in Table 5.

| Hyperparameter | Value |
|---|---|
| Number of Objectives | 3 |
| Population Size | 50 |
| Initial Species Size | 10 |
| Number of Uniform Reference Points | 25 |
| UCB Coefficient | 0.9 |
| Crossover Probability | 1.0 |
| Mutation Operation Probability | 1.0 |
| Neuroevolution Mutation Fraction | 0.05 |
| Neuroevolution Mutation Strength | 0.1 |

**Table 4: NEMO Search Hyperparameters**

In our study, NEMO has four distinct species: two direct search methods and two GNN based species. The architecture design of GNNs follows the sequence:

(1) Graph Layers: (Graph-Conv Kipf and Welling [23] or Graph-UNet [12])

(2) SELU Activation Function
(3) Graph Attention Layer [43]
(4) SELU Activation Function
(5) Linear Layer
(6) SELU Activation Function

Table 5 shows the hyperparameters for the GNN species, Graph-Conv and Graph-UNet, which define the neural network architecture for each species, and can be modified depending on the underlying optimization problem.

| Hyperparameter | Value |
|---|---|
| Graph-UNet Depth | 3 |
| Graph-UNet Attention Heads | 4 |
| Graph-UNet Hidden Sizes | 10, 8 |
| Graph-Conv-Net Attention Heads | 4 |
| Graph-Conv-Net Hidden Sizes | 10, 8 |

**Table 5: NEMO Species Hyperparameters**

### B.2 Graph Embedding

A given neural network workload is first transformed to a quantized model via NNCF[25] as described inSection 3.1, where NNCF automatically determines and inserts quantization operation at the quantizable weight and activation nodes. Table 6 describes the features included in each node of our graph representation of the quantized model shown schematically in Figure 1 and Figure 2. While HAQ[44] creates a tailored embedding for a set of network layers per convolutional network, we design 5 basic features below to generalize node representation for any neural network operations.

| Node Features | Description |
|---|---|
| $Op_{id}$ | One-hot encoded node operation type |
| $W_{bool}$ | True for weight tensor; false for activation tensor |
| $T_{ndim}$ | number of quantizable tensor dimensions |
| $T_{numel}$ | number of elements in a quantizable tensor |
| $G_{id}$ | One-hot encoded group ID dependent on the available groups of quantizable tensors |

**Table 6: Graph Node Features**